\PassOptionsToPackage{table}{xcolor}
\documentclass{bmvc2k}
\usepackage{booktabs}
\usepackage{pifont}
\usepackage{multirow}
\usepackage{array}
\usepackage{amsmath}
\usepackage{amssymb}

\definecolor{oursrow}{RGB}{236,245,255}
\definecolor{promptblue}{RGB}{25,80,180}
\definecolor{promptorange}{RGB}{194,88,19}
\definecolor{promptbg}{RGB}{248,250,252}
\definecolor{promptborder}{RGB}{148,163,184}
\definecolor{promptlabel}{RGB}{51,65,85}
\definecolor{tickgreen}{RGB}{31,140,64}
\definecolor{crossred}{RGB}{192,38,38}
\definecolor{swapcell}{RGB}{255,241,219}
\newcommand{\cmark}{\textcolor{tickgreen}{\ding{51}}}
\newcommand{\xmark}{\textcolor{crossred}{\ding{55}}}
\newcommand{\dzero}{\textcolor{promptlabel}{\normalfont\tiny(0.0)}}
\newcommand{\dpos}[1]{\textcolor{tickgreen}{\normalfont\tiny(+#1)}}
\newcommand{\dneg}[1]{\textcolor{crossred}{\normalfont\tiny(-#1)}}
\newcommand{\swapres}[1]{\cellcolor{swapcell}#1}
\newcommand{\addstage}[1]{\textcolor{tickgreen}{\textbf{\ding{58}}}\,#1}
\newcommand{\remstage}[1]{\textcolor{crossred}{\textbf{\ding{54}}}\,#1}
\newcommand{\chatbos}{\textcolor{promptlabel}{\texttt{\textless|begin\_of\_text|\textgreater}}}
\newcommand{\chateot}{\textcolor{promptlabel}{\texttt{\textless|eot\_id|\textgreater}}}
\newcommand{\chatdump}[1]{\noindent{\footnotesize\raggedright #1\par}}
\newcommand{\systempromptref}{\textcolor{promptlabel}{\texttt{[system prompt]}}}
\newcommand{\promptcard}[3]{%
    \par\noindent\begingroup
    \setlength{\fboxsep}{6pt}%
    \fcolorbox{promptborder}{promptbg}{%
        \begin{minipage}{0.93\linewidth}
        \small\raggedright
        \textbf{#1}\hfill{\footnotesize\textcolor{promptlabel}{#2}}\par
        \vspace{0.25em}{\color{promptborder}\hrule}\vspace{0.45em}
        #3
        \end{minipage}}%
    \endgroup\par\vspace{0.35em}%
}

\newcommand{\vtok}{\texttt{\textless v\textgreater}}
\newcommand{\assistantspan}[1]{\textcolor{promptorange}{\textit{#1}}}
\newcommand{\mote}{MoTE}
\newcommand{\motefull}{Mixture of Task Experts (MoTE)}
\newcommand{\videollmmote}{VideoLLM-\allowbreak MoTE}
\newcommand{\motemodel}[1]{\videollmmote-\allowbreak 1B+\allowbreak #1E}
\newcommand{\glmocrmote}{GLM-OCR-\allowbreak MoTE-\allowbreak 0.9B+\allowbreak 2E}
\title{MoTE: Mixture of Task Experts for Multi-Task Video Understanding}

\addauthor{Muhammad Asad Ali$^{*}$}{muhammad_asad.ali@dfki.de}{1,2}
\addauthor{Umar Khan$^{*}$}{umaryousafzai9@gmail.com}{1}
\addauthor{Nadia Robertini}{nadia.robertini@dfki.de}{2}
\addauthor{Didier Stricker}{didier.stricker@dfki.de}{1,2}
\addinstitution{
 Department of Computer Science\\
 University of Kaiserslautern-Landau (RPTU)\\
 Kaiserslautern, Germany
}
\addinstitution{
 Augmented Vision Group \\
 German Research Center for Artificial Intelligence (DFKI)\\
 Kaiserslautern, Germany
}

\runninghead{Ali, Khan, Robertini, Stricker}{Mixture of Task Experts}

\begin{document}

\begingroup
\renewcommand{\thefootnote}{*}
\footnotetext{Equal contribution.}
\endgroup

\maketitle

\begin{abstract}
\label{sec:abstract}
Procedural video-language models must solve heterogeneous tasks from the same visual evidence, including action recognition, forecasting, and procedure prediction. Dense transformer decoders share the same feed-forward networks across tasks, which can entangle task behavior and make controlled capability expansion difficult. Sparse Mixture-of-Experts (MoE) decoders provide conditional computation, but token-level learned routing is not naturally aligned with task-level procedural objectives. We propose \textbf{\motefull}, a decoder architecture that converts large language model feed-forward networks into task-specific experts while keeping the multimodal backbone shared. Each example follows one sample-level task route, so active task-expert computation remains independent of the number of stored task experts. We instantiate this design as \textbf{\videollmmote} and evaluate it on five COIN benchmarks using explicit task routes. The five-expert model activates \textasciitilde2B LLM parameters per sample and achieves higher average top-1 accuracy than recent VideoLLM baselines. Under the same expert topology, it improves over dense all-expert activation and learned sparse-routing controls. These results show that task-structured routing provides an interpretable and compute-efficient decoder alternative for multi-task video-language learning. 
\end{abstract}
%-------------------------------------------------------------------------
\section{Introduction}
\label{sec:intro}

Intelligent multimodal assistants are expected to follow human instructions over visual observations, from answering what is happening in a procedure to anticipating the next action or producing a concise procedural summary. Instruction tuning has become an effective way to align vision-language models with such diverse user requests, connecting pretrained visual encoders to language models through learned projectors and instruction-following supervision~\cite{li2023blip2,liu2023llava,dai2023instructblip}. Video-language models extend this interface to temporal inputs, where a deployed assistant may receive either a complete clip or a live stream of frames~\cite{joya2024videollmonline,dibyadip2025memoryefficient,rui2025dispider,junming2024streamingbench,zhenpeng2025online}. In procedural video understanding, however, the challenge is not only to encode time. The same visual evidence must support heterogeneous tasks such as recognizing the current action, forecasting future steps, identifying the goal, and producing procedure-level responses.

This task diversity creates a concrete modeling tension. Multi-task instruction tuning can improve generalization by exposing a model to many task formats~\cite{sanh2022t0,dai2023instructblip}, but prior vision-language studies also show that mixing dissimilar tasks can create negative transfer when all tasks update the same parameters~\cite{gou2024mocle}. For procedural videos, task prompts often differ in temporal horizon and output structure even when the underlying frames are shared. A dense decoder therefore has to encode task-specific transformations inside the same feed-forward blocks used by every task.

\begin{figure}[!t]
    \centering
    \includegraphics[
    width=\columnwidth,
    trim={0cm 12cm 0cm 0cm},
    clip
    ]{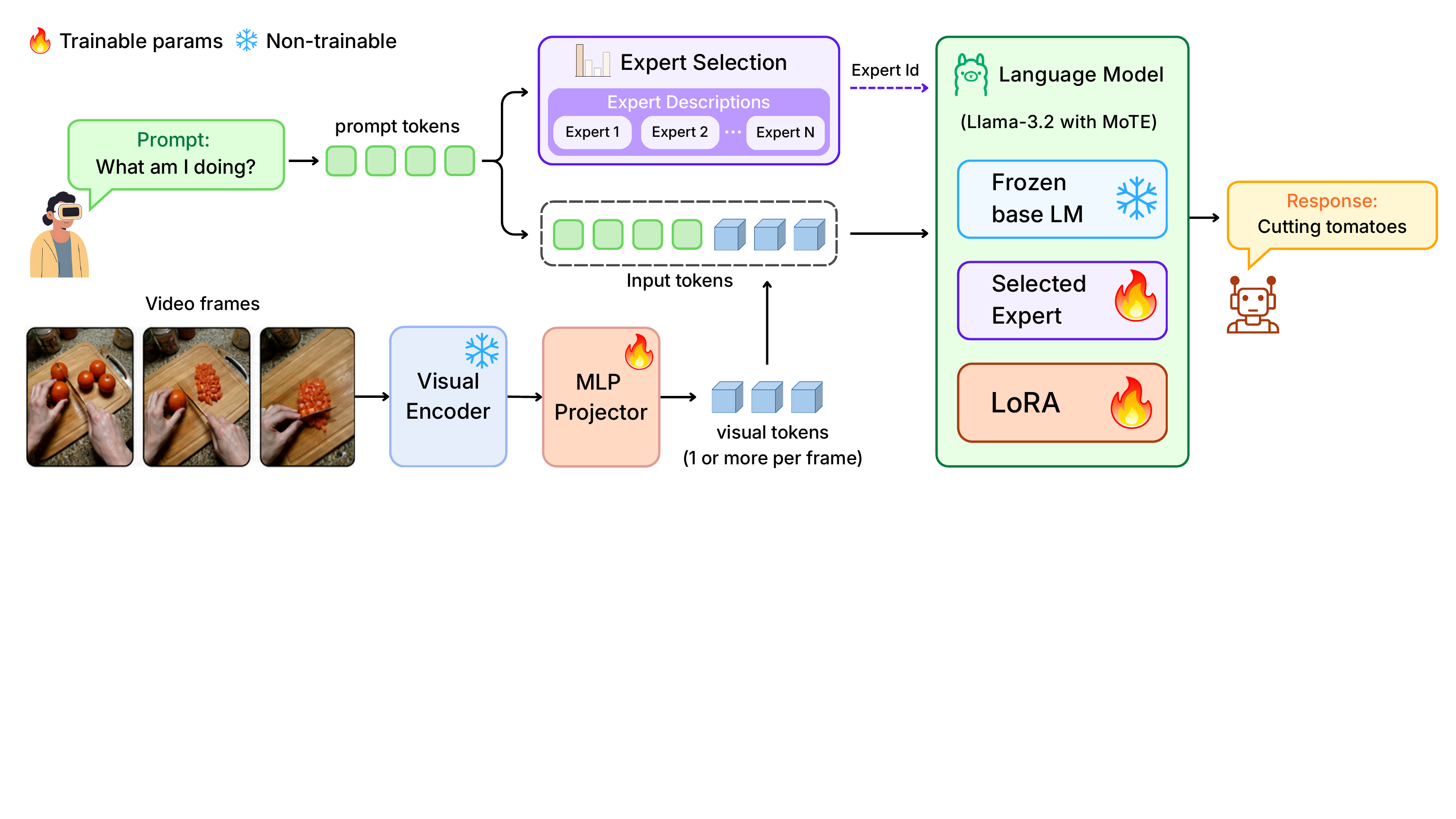}
    \caption{\textbf{\videollmmote{} overview.} A user prompt and sampled video frames are converted into prompt and visual tokens before entering an LLM decoder modified with \mote{}. Prompt-conditioned expert selection matches the prompt to registered expert descriptions and passes the selected expert id to the decoder. Given the input tokens and expert id, LLM generates a response. The visual encoder and base language model are non-trainable, while the MLP projector, LoRA adapters, and selected task experts are trainable. }
    \label{fig:mote_full}
\end{figure}

Modern video-language systems have made progress on long and streaming inputs by improving memory, causal processing, token selection, and computation scheduling~\cite{joya2024videollmonline,enhong2024videollmmod,dibyadip2025memoryefficient,guo2026eventvstream}. These advances make video inputs more usable, but decoder-side task specialization often remains implicit: the same language-decoder path must express every prompt type. A useful multi-task video-language decoder should therefore preserve shared visual-language grounding while exposing task-specific computation in a controllable part of the model.

Sparse Mixture-of-Experts (MoE) transformers offer a natural starting point for conditional computation~\cite{shazeer2017outrageously,fedus2021switch,riquelme2021scalingvision}. In language, vision, and large vision-language model settings, sparse experts can increase stored capacity while activating only a subset of parameters per input~\cite{lepikhin2020gshard,lin2024moellava}. Standard sparse MoE layers, however, usually route tokens or patches with learned gates. That granularity is effective for scaling, but it does not expose a stable task-level computation path for procedural prompts. Our setting uses known procedural task structure and asks whether the decoder itself can expose task-level routes.

We propose \textbf{\motefull}, a decoder architecture that makes task identity an explicit routing variable. \mote{} converts each decoder feed-forward network (FFN) into task-specific experts and an always-active shared expert, while keeping the video-language backbone and attention path shared. Each video--prompt example follows one sample-level task route, so active task computation does not grow with the number of stored task experts. Because experts are explicit modules, the architecture can adapt to new tasks by adding task experts while retaining previously learned knowledge in the shared backbone and existing experts. This serves as an architectural principle that we demonstrate is applicable across different domains. During training, task labels provide direct route supervision; at inference time, the route can be supplied externally or selected by matching the prompt to registered expert descriptions.

We evaluate this architecture as \videollmmote{} on five COIN tasks. VideoLLM-MoTE outperforms recent COIN baselines while using a smaller active-parameter budget. To verify the cross-domain applicability of this architecture, we further conduct tests on document understanding, successfully demonstrating task separation for Key Information Extraction (KIE) and Optical Character Recognition (OCR) across two different datasets. Expert-addition analysis and the explicit expert-removal mechanism further illustrate how task-aligned FFN experts can serve as modular units.

Our contributions are:
\begin{enumerate}
    \item We introduce \mote{}, a modular decoder conversion that replaces shared LLM feed-forward networks with task-routed FFN experts, while retaining shared components for task-agnostic transformations.
    
    \item We instantiate \mote{} as \videollmmote{} for multi-task procedural video understanding and show that task-level routing improves over recent VideoLLM baselines, and learned sparse-routing controls under the same expert topology.
    
    \item We show that the same decoder expertization principle transfers to document analysis domain by converting GLM-OCR~\cite{glmocr2026} into GLM-OCR-MoTE, separating OCR and receipt key information extraction (KIE) task routes.
    
    \item We analyze the modularity of \mote{} through expert addition and removal, showing that task experts can be introduced or removed as architectural modules while preserving previously learned knowledge.
\end{enumerate}

%-------------------------------------------------------------------------
\section{Related Work}
\label{sec:related_work}
\paragraph{Streaming Procedural Video Understanding.}
Recent video-language work moves from offline video understanding toward causal, deployable assistants that must process frames as they arrive. VideoLLM-online~\cite{joya2024videollmonline} formalizes streaming dialogue, where a model must decide when to answer during an incoming video stream. VideoLLM-MoD~\cite{enhong2024videollmmod} reduces redundant layer computation with mixture-of-depths routing while retaining richer visual tokens. Other systems improve long-context memory, causal temporal modeling, event-triggered reasoning, and always-on interaction~\cite{dibyadip2025memoryefficient,yibin2025learning,guo2026eventvstream,guan2026videostreamingthinking,lu2026aura,wen2026evostreaming}. Recent benchmarks further show that online video understanding requires multiple procedural capabilities, including timestamped perception, anticipation, temporal reasoning, and multi-turn assessment~\cite{tang2026streamingeval,junming2024streamingbench,junbo2025ovobench,mu2024temporalbench,zhenyu2025svbench,zhenpeng2025online}. These requirements make procedural video-language modeling inherently multi-task. Existing streaming methods mainly improve memory, token selection, visual encoding, or computation depth. In contrast, \mote{} studies how task-specific decoder computation can support heterogeneous procedural objectives while keeping the video-language backbone shared.

\paragraph{Instruction Tuning and Task Conflict.}
Instruction-tuned vision-language models connect pretrained visual encoders to large language models through lightweight projectors, enabling general-purpose systems such as BLIP-2, LLaVA, and InstructBLIP~\cite{li2023blip2,liu2023llava,dai2023instructblip}. Instruction tuning teaches these models to map diverse task templates to appropriate textual responses~\cite{sanh2022t0}, and visual instruction tuning can improve held-out task generalization~\cite{dai2023instructblip}. However, this shared-decoder design also creates a multi-task learning problem: heterogeneous prompts, objectives, and output formats compete for the same model capacity. This conflict is well studied in multi-task learning, where shared representations can improve transfer but may also cause negative transfer when task objectives interfere~\cite{ruder2017overviewmultitasklearningdeep,vandenhende2021multitasklearningdenseprediction,chen2018gradnormgradientnormalizationadaptive,sener2018multitasklearningmultiobjectiveoptimization,yu2020gradientsurgerymultitasklearning}. In procedural video-language modeling, this conflict is sharper because the same visual evidence supports recognition, forecasting, and procedure-level reasoning, yet each task requires different temporal decisions and output formats. \mote{} targets this setting by keeping the video-language backbone shared while assigning task-specific transformations to decoder FFN experts.

\paragraph{Sparse MoE and Task-Level Routing.}
Sparse Mixture-of-Experts transformers add conditional capacity through routed experts. Foundational MoE systems route tokens or image patches with learned gates for language and vision scaling~\cite{shazeer2017outrageously,lepikhin2020gshard,fedus2021switch,riquelme2021scalingvision}, while dense-to-sparse construction reduces the cost of building MoE models by copying or partitioning pretrained FFNs into experts~\cite{komatsuzaki2022sparseupcycling,zhu2024llamamoe,lin2024moellava}. These designs scale capacity effectively, but token- or patch-level gates are not naturally aligned with user-facing task boundaries. Task-aware MoE methods move routing closer to task semantics by using task indices, task representations, or task-conditioned features in the routing function. Cross-task MoE uses task representations to select layer-wise expert mixtures across NLP tasks~\cite{ye2022crosstaskmoe}, M$^3$ViT uses task-dependent routing for multi-task vision~\cite{liang2022m3vit}, and Edge-MoE focuses on efficient execution for task-aware ViT MoE models~\cite{sarkar2023edgemoe}. MoCLE addresses task conflict in instruction tuning with LoRA experts conditioned on instruction clusters~\cite{gou2024mocle,hu2021lora}. \mote{} differs by placing specialization inside the LLM decoder FFN path of a video-language model: each video--prompt sample selects one task route, activating the corresponding FFN expert at each decoder layer while attention, the visual encoder, and the MLP projector remain shared. Table~\ref{tab:moe_comparison} summarizes these distinctions across routing level, routing type, interpretability, and task awareness.

\begin{table}[t]
\centering
\small
\resizebox{\textwidth}{!}{%
\begin{tabular}{l|c|c|c|c|c}
\toprule
Method & Domain & Routing Level & Routing Type & Interpretability & Task-Aware \\
\midrule
Sparse MoE~\cite{shazeer2017outrageously} & Language & Token & Learned top-$k$ & Low & No \\
GShard~\cite{lepikhin2020gshard} & Language & Token & Learned top-$k$ & Low & No \\
Switch Transformer~\cite{fedus2021switch} & Language & Token & Learned top-1 & Low & No \\
V-MoE~\cite{riquelme2021scalingvision} & Vision & Patch & Learned top-$k$ & Low & No \\
Sparse Upcycling~\cite{komatsuzaki2022sparseupcycling} & Language/vision & Token & Expert Choice/Learned top-$k$ & Low & No \\
LLaMA-MoE~\cite{zhu2024llamamoe} & Language & Token & Learned top-$k$ & Low & No \\
MoE-LLaVA~\cite{lin2024moellava} & Vision-language & Token & Learned top-$k$ & Low & No \\
MoCLE~\cite{gou2024mocle} & LVLM instruction & Sample & Cluster-conditioned learned top-1 & Medium & Yes \\
M$^3$ViT~\cite{liang2022m3vit} & Multitask vision & Token/Patch & Task-conditioned learned top-$k$ & Medium & Yes \\
Edge-MoE~\cite{sarkar2023edgemoe} & Multitask vision & Token/Patch & Task-conditioned learned top-$k$ & Medium & Yes \\
Cross-task MoE~\cite{ye2022crosstaskmoe} & Language & Sample & Task-conditioned learned top-$k$ & Medium & Yes \\
\midrule
\textbf{Ours (\mote)} & Video-language & Sample & Task index / prompt-conditioned expert selection & High & Yes \\
\bottomrule
\end{tabular}%
}
\caption{
Comparison of MoE approaches. Prior sparse transformers primarily route tokens or patches with learned gates. Task-level methods make routing more aligned with task semantics by using task-conditioned features in the gating function, but they are not designed for modular video-language decoders. \mote{} uses sample-level task routing over decoder FFN experts; the reported experiments use explicit task routes, while prompt-description matching provides an expert-selection component for inference. 
}
\label{tab:moe_comparison}
\end{table}

%-------------------------------------------------------------------------
\section{Method}
\label{sec:method}

\subsection{Overview}
\label{sec:method_overview}

We study multi-task video-language learning when the training set provides a discrete task structure. Each example is a tuple $(v, x, t, y)$, where $v$ is a video or sampled frame sequence, $x$ is a natural-language prompt, $t \in \{1,\dots,M\}$ is the task identity, and $y$ is the target response. The learning objective is to model $p_\theta(y \mid v, x, t)$ while retaining the ability to answer at inference time when the task is either explicitly provided or only implied by the prompt. 

As shown in Figure~\ref{fig:mote_full}, \videollmmote{} consists of a vision encoder, an MLP projector, and a large language model with \mote{} FFN expert blocks. We introduce the architecture of \videollmmote{} in Section~\ref{sec:method_architecture}, the training objective in Section~\ref{sec:method_training}, and modular task expert adaptation in Section~\ref{sec:method_modularity}.

\subsection{Architecture of \videollmmote}
\label{sec:method_architecture}

\paragraph{Shared Backbone.}
\label{sec:method_backbone}
The shared backbone follows the standard video-language pattern shown in Figure~\ref{fig:mote_full}. A pretrained vision encoder maps the sampled frames in $v$ to a visual token sequence $Z=[z_1,\dots,z_P]\in\mathbb{R}^{P\times C}$, where $P$ is the total number of visual tokens and $C$ is the visual feature dimension. A trainable two-layer MLP projector $f_{\mathrm{proj}}$ maps these features into the LLM hidden size $D$, giving $U_v=f_{\mathrm{proj}}(Z)\in\mathbb{R}^{P\times D}$. The prompt $x$ is embedded by the language model token embedding layer, yielding $U_x\in\mathbb{R}^{N\times D}$ for $N$ text tokens. The chat template marks visual positions with \texttt{<v>}; after projection, those placeholders are replaced by $U_v$, producing the decoder input sequence $H_0=\mathrm{Template}(U_x,U_v)$, which concatenates visual and text token embeddings. All samples share this visual encoder, MLP projector, language token embedding layer, and attention path. Further implementation details such as frame rate, resolution, visual-token count, and backbone scale are reported in Section~\ref{sec:experimental_setup}.

\begin{figure}[!t]
    \centering
    \includegraphics[width=\columnwidth]{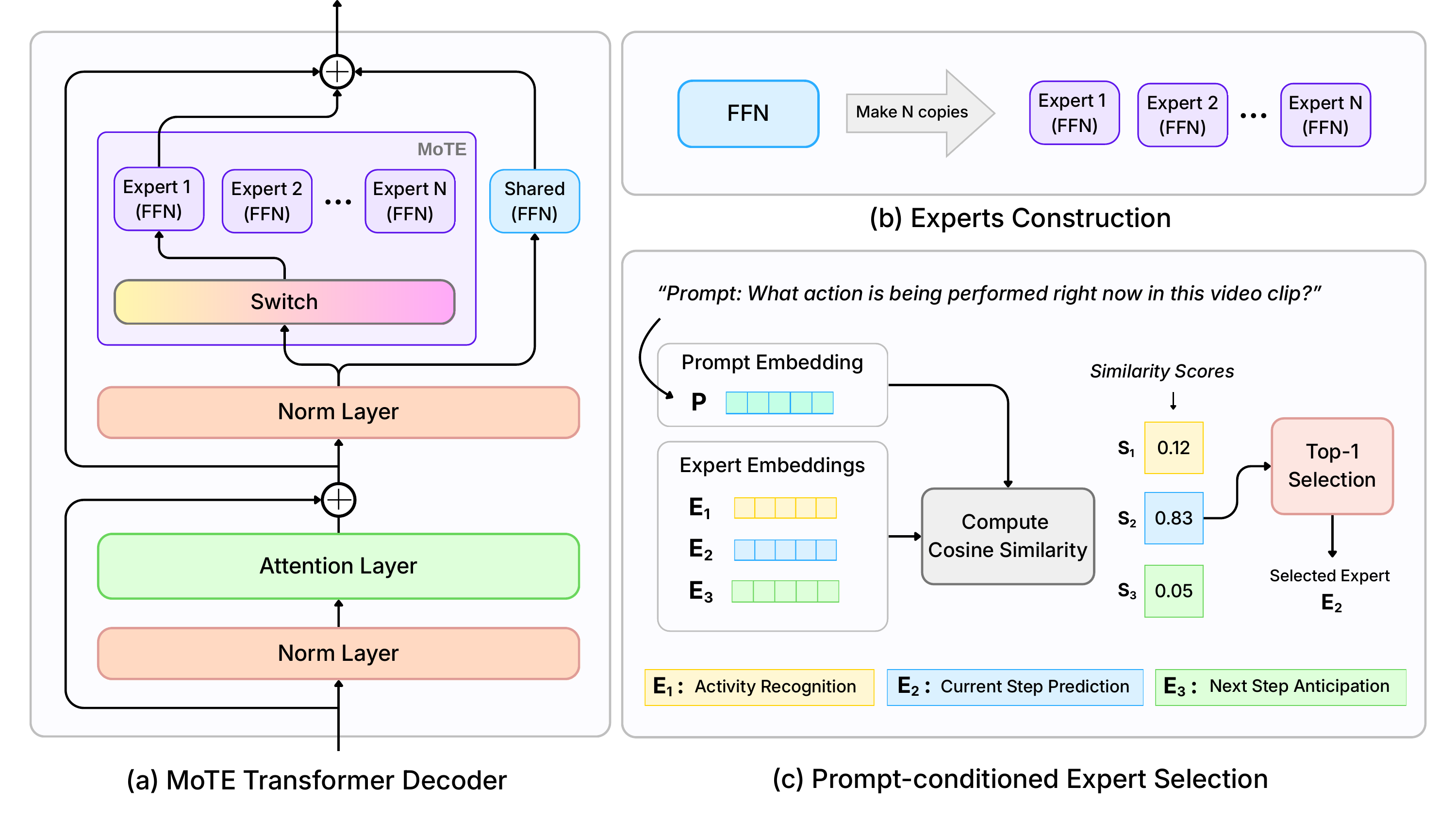}
    \caption{\textbf{\mote{} decoder and expert selection.} (a) In each decoder layer, attention remains shared, while the FFN layer is replaced by a shared FFN expert and a switch-selected task FFN experts. Both expert outputs are added to the residual stream. (b) Task experts are initialized by copying the pretrained dense FFN into multiple task branches. (c) Prompt-conditioned expert selection embeds the prompt and registered expert descriptions, computes cosine similarities, and selects the top-1 expert for the full video--prompt sample.}
    \label{fig:mote_comps}
\end{figure}

\paragraph{MoTE decoder blocks.}
\label{sec:method_task_experts}
\mote{} modifies the transformer decoder layer in the LLM. Only the FFN sublayer is changed, so task-dependent transformations can be selected without duplicating the full video-language model. Concretely, let the dense decoder contain $L$ layers. In layer $l$, the original FFN is denoted by $F_l$. As shown in Figure~\ref{fig:mote_comps}(a), \mote{} replaces $F_l$ with a set of task experts
\[
\mathcal{F}_l = \{F_l^1, F_l^2, \dots, F_l^M\}
\]
and a shared expert $F_l^{\mathrm{sh}}$. Each task expert has the same architecture as the original FFN and is associated with one task. The shared expert is active for every task and captures transformations that should remain common, while the selected task expert specializes to task-specific supervision and output formats.

For hidden states $H_l$, the shared pre-norm attention block produces
\[
\widetilde{H}_l = H_l + \mathrm{GQA}_l(\mathrm{LN}(H_l)),
\]
where $\mathrm{GQA}_l$ denotes grouped-query attention and $\mathrm{LN}(\cdot)$ denotes layer normalization. Given a task route $r \in \{1,\dots,M\}$, the expert block computes
\[
H_{l+1}
= \widetilde{H}_l
+ \lambda_{\mathrm{sh}} F_l^{\mathrm{sh}}(\mathrm{LN}(\widetilde{H}_l))
+ \lambda_{\mathrm{task}} F_l^{r}(\mathrm{LN}(\widetilde{H}_l)).
\]
The coefficients $\lambda_{\mathrm{sh}}$ and $\lambda_{\mathrm{task}}$ are expert scaling factors; in our runs the routed scaling factor is 1.0. As illustrated in Figure~\ref{fig:mote_comps}(b), both routed and shared experts are initialized as copies of the pretrained dense FFN, following the dense-to-expert initialization principle used in sparse upcycling~\cite{komatsuzaki2022sparseupcycling}. This avoids starting task experts from random transformations and allows the shared and task-specific paths to diverge during training.

This construction separates two roles that are entangled in a dense decoder. The shared expert learns common video-language transformations, while $F_l^m$ can specialize to the supervision and output format of task $m$. Because only one task expert is activated for a sample, the number of stored experts can grow with the task set without requiring all experts to run for every input.

% Let $T$ denote the number of routed task experts in a VideoLLM instantiation, excluding the always-active shared expert. We name each variant as \textit{OriginalModel}-\allowbreak MoTE-\allowbreak \textit{BaseSize}+$T$E, where the base-size term denotes the underlying LLM scale and $T$E counts only the routed task experts. For example, \motemodel{5} applies \mote{} to a 1B VideoLLM backbone with five routed task experts plus the shared expert. We report parameter counts separately whenever model size is discussed. Thus, the COIN model is \motemodel{5}, and the COIN+Ego4D expanded model is \motemodel{6}.

\paragraph{Task-route selection.}
\label{sec:method_routing}

\mote{} routes the full video--prompt sample rather than individual tokens: all visual and textual tokens in an example pass through the same task expert at every decoder layer. This aligns computation with the user-facing task and avoids a learned-gate or load-balancing loss in the explicit task-index setting. For a training tuple $(v,x,t,y)$, where $v$ is the video input, $x$ is the prompt, and $t \in \{1,\dots,M\}$ is the task identity among $M$ task experts, the direct route is $r_{\mathrm{idx}}(v,x,t)=t$.

During joint training, this known task identity provides route supervision. Only the selected task expert and the shared expert receive gradients for a given example, exposing each task expert to a consistent task distribution.

At inference time, the task identity may be unavailable. As illustrated in Figure~\ref{fig:mote_comps}(c), each expert index $m \in \{1,\dots,M\}$ is associated with a natural-language expert description $d_m$, such as ``predict the next step'' or ``identify the overall procedure''. Let $\phi(\cdot)$ be a frozen text encoder, and let $\bar{\phi}(z)=\phi(z)/\|\phi(z)\|_2$ denote the normalized embedding of any text string $z$. We infer a prompt-conditioned route by nearest-neighbor matching between the prompt and the expert descriptions:
\[
r_{\mathrm{prompt}}(x)=\arg\max_{m \in \{1,\dots,M\}}
\bar{\phi}(x)^\top \bar{\phi}(d_m).
\]
This selector is useful for modular deployment. In contrast to a learned gate whose output dimension is fixed by the experts observed during training, an expert description registry can add or remove natural-language expert descriptions while leaving the decoder parameters and remaining expert routes unchanged. We therefore use prompt-conditioned selection as an architectural component for inference when a task index is unavailable.

\subsection{Training Objective}
\label{sec:method_training}

Following VideoLLM-online~\cite{joya2024videollmonline}, the training objective supervises two behaviors in the same autoregressive video-language sequence. At response tokens, the model predicts the next answer token under the selected expert route. At streaming positions that don't require a response, it predicts EOS so it stays silent until needed. Let $p_\theta(\cdot \mid H_{\leq j}, r)$ denote the next-token distribution after prefix $H_{\leq j}$ under route $r$. We use two binary indicators: $l_j=1$ when position $j$ is a supervised language-response token, and $s_j=1$ when position $j$ is a frame position assigned streaming silence supervision. The loss is
\[
\mathcal{L}
= \frac{1}{N}\sum_{j=1}^{N}
(\underbrace{-l_{j+1}\log p_\theta(a_{j+1}\mid H_{\leq j},r)}_{\text{LM Loss}} - \underbrace{w s_j\log p_\theta(\mathrm{EOS}\mid H_{\leq j},r)}_{\text{Stream Loss}})
,
\]
where $a_{j+1}$ is the next target token, $w$ balances the streaming term, and $N$ is the sequence length. We use $w=1$ in reported runs. In our implementation, $s_j$ is nonzero only for the last visual token of a frame when that frame is not immediately followed by a supervised language response. For all COIN offline benchmark tasks, stream messages are marked as non-learned positions, so $s_j=0$ for frame tokens and the loss reduces to standard routed language modeling:
\[
\mathcal{L}_{\mathrm{COIN}}
= -\frac{1}{N}\sum_{j=1}^{N}
l_{j+1}\log p_\theta(a_{j+1}\mid H_{\leq j},r).
\]
% For Ego4D narration, we keep the streaming EOS term, so the narration expert learns both what to say and when to remain silent.

\begin{table*}[t]
\centering
\small
\setlength{\tabcolsep}{3.5pt}
\resizebox{0.9\textwidth}{!}{%
\begin{tabular}{@{}l|c|ccccc|c@{}}
\toprule
\multirow{2}{*}{Method} &
\multirow{2}{*}{\shortstack{Not use\\HowTo100M}} &
\multicolumn{5}{c|}{COIN Benchmark Top-1 Accuracy$\uparrow$} &
\multirow{2}{*}{Avg.} \\
& & Step & Next & Proc. & Task & Proc.+ & \\
\midrule
\multicolumn{8}{@{}l}{\textcolor{gray}{\footnotesize\textit{Non-LLM procedural video methods}}} \\
ClipBERT~\cite{lei2021clipbert} & \cmark & 30.8 & -- & -- & 65.4 & -- & -- \\
TimeSformer~\cite{bertasius2021timesformer} & \xmark & 46.5 & 34.0 & 17.0 & 85.3 & 40.1 & 44.6 \\
Paprika~\cite{zhou2023paprika} & \xmark & 51.0 & 43.2 & -- & 85.8 & -- & -- \\
DistantSup~\cite{lin2022distant} & \xmark & 54.1 & 39.4 & -- & 90.0 & 41.3 & -- \\
VideoTF~\cite{narasimhan2023learning} & \xmark & 56.5 & 42.4 & 40.2 & 91.0 & 46.4 & 55.3 \\
ProcedureVRL~\cite{zhong2023learning} & \xmark & 56.9 & 46.8 & -- & 90.8 & -- & -- \\
VideoTaskGraph~\cite{ashutosh2024videomined} & \xmark & 57.2 & 40.2 & -- & 90.5 & -- & -- \\
\midrule
\multicolumn{8}{@{}l}{\textcolor{gray}{\footnotesize\textit{LLM-based video-language methods}}} \\
VideoLLM-online-7B-v1~\cite{joya2024videollmonline} & \cmark & 59.8 & 48.1 & 47.9 & 92.1 & 52.9 & 60.2 \\
VideoLLM-online-8B-v1+~\cite{joya2024videollmonline} & \cmark & 63.1 & 49.1 & 49.8 & 92.7 & 54.1 & 61.8 \\
VideoLLM-MoD~\cite{enhong2024videollmmod} & \cmark & 63.4 & 49.7 & 49.8 & 92.8 & 53.3 & 61.8 \\
\rowcolor{oursrow}VideoLLM-online-1B-v1+ & \cmark & 57.7 & 47.4 & 48.1 & 92.1 & 50.8 & 59.3 \\
\rowcolor{oursrow}VideoLLM-online-3B-v1+ & \cmark & 59.5 & 47.9 & 48.3 & 92.4 & 51.4 & 59.9 \\
\rowcolor{oursrow}\textbf{\motemodel{5} (ours)} & \cmark & \textbf{65.1} & \textbf{50.3} & \textbf{50.5} & \textbf{94.4} & \textbf{54.2} & \textbf{62.9} \\
\bottomrule
\end{tabular}
}
\caption{
Comparison with prior COIN baselines. \textit{Not use HowTo100M} indicates whether the method avoids HowTo100M pretraining. The average is reported when all five task scores are available. Our method achieves 62.9\% average accuracy outperforming other baselines.
}
\label{tab:coin_public_baselines}
\end{table*}

\subsection{Modular Task Expert Adaptation}
\label{sec:method_modularity}

\mote{} supports task adaptation by changing the task-specific FFN experts associated with the affected capability. New tasks can be introduced by attaching new routed experts, while tasks that are no longer needed can be removed from the active expert set without rewriting the shared video-language backbone or the remaining task experts.

\paragraph{Expert addition.}
\label{sec:method_addition}
Assume a model has been trained for tasks $\{1,\dots,K\}$. To add task $K+1$, we insert a new expert $F_l^{K+1}$ into every decoder layer and initialize it from the corresponding dense FFN or the shared expert. We then train on the new task using route $r=K+1$ while freezing the visual encoder, MLP projector, attention blocks, shared experts, and existing task experts. Only the new experts are updated:
\[
\Theta_{\mathrm{train}}
= \{\theta(F_l^{K+1}) : l=1,\dots,L\}.
\]
This isolates new-task learning from previously learned task modules: old routes keep the parameters that produced their earlier behavior, which helps mitigate catastrophic forgetting during expansion and prevents knowledge overwriting. 

\paragraph{Expert removal.}
\label{sec:method_removal}
When a task is outside the target deployment domain, or when an expert capability should be withdrawn (i.e. for compliance reasons), its experts can be dropped from every decoder layer and its description can be removed from the selector. This gives \mote{} a direct capability-removal mechanism that is difficult to express in a dense decoder, where task behavior is distributed across shared parameters. 

%-------------------------------------------------------------------------
\section{Experiments}
\label{sec:experiments}

\subsection{Experimental Setup}
\label{sec:experimental_setup}

\paragraph{Datasets and tasks.}
Following VideoLLM-online~\cite{joya2024videollmonline} and VideoLLM-MoD~\cite{enhong2024videollmmod}, we evaluate \videollmmote{} in both offline procedural benchmarks and an online narration setting. For offline evaluation, we use COIN~\cite{Tang_2019_CVPR}, an instructional-video dataset with temporally annotated procedural steps. We report five COIN benchmark tasks: \textit{Step} recognition, \textit{Next} step forecasting, \textit{Task} recognition, \textit{Proc.} procedure forecasting, and \textit{Proc.+} task-conditioned procedure forecasting. We use \motemodel{5} for this setting, with one routed expert for each COIN task. To evaluate task extension across datasets, we further use the Ego4D narration stream~\cite{Grauman_2022_CVPR}, where timestamped egocentric narrations define when the model should speak and what it should say. We first train an Ego4D narration route and then attach five COIN routes trained on the COIN tasks. The resulting \motemodel{6} is evaluated as a single expanded model that preserves the original online narration capability while adding offline procedural reasoning tasks. Finally, we test whether the same MoTE method transfers beyond video understanding. We convert GLM-OCR-0.9B into \glmocrmote{} by retaining an OCR expert and adding a receipt key information extraction (KIE) expert. We evaluate direct image-to-structured-field extraction on SROIE~\cite{huang2019icdar2019} and CORD~\cite{park2019cord} datasets, while also measuring whether the original OCR route is preserved after adding the KIE route.

\paragraph{Implementation details.}
All reported \videollmmote{} models are trained on four A100-80GB GPUs with videos sampled at 2 FPS and capped at 1200 frames. Our default COIN setting uses a SigLIP 2 ViT-L/16-384 visual encoder~\cite{tschannen2025siglip2multilingualvisionlanguage}, a two-layer MLP projector, and a Llama-3.2-1B-Instruct decoder~\cite{grattafiori2024llama3herdmodels}. Given our limited computational resources, we select this 1B decoder as the default backbone rather than training larger variants; this keeps the study focused on routing behavior under a tractable training budget. Each frame is represented by one CLS token plus a $3{\times}3$ pooled spatial grid, giving 10 visual tokens per frame. We fine-tune the language model with LoRA~\cite{hu2021lora} and keep the MLP projector trainable. The COIN model contains five routed task experts and one shared expert; each sample activates only the selected routed expert and the shared expert. Supplementary sections ~\ref{sec:supp_architecture_params}--~\ref{sec:supp_bucketed_sampling} provide details about model parameters, training configuration and data sampling technique.

\paragraph{Evaluation Metrics.}
For COIN, we follow the generated-answer evaluation used by prior procedural video-language work~\cite{joya2024videollmonline,enhong2024videollmmod}: model outputs are decoded as text, normalized, mapped back to the task label space, and scored as top-1 accuracy. For \textit{Proc.} and \textit{Proc.+}, the generated numbered list is evaluated step-wise over the aligned future-action positions. We report the arithmetic mean over the five COIN tasks when all task scores are available. For Ego4D narration validation, following VideoLLM-online, we report perplexity (PPL) for language modeling quality, TimeDiff for temporal response alignment, and Fluency as a combined language-and-timing score. Lower is better for PPL and TimeDiff, while higher is better for Fluency. We refer readers to~\cite{joya2024videollmonline} for the detailed definition and computation of the online narration metrics.

\begin{table*}[t]
\centering
\small
\setlength{\tabcolsep}{3pt}
\resizebox{0.9\textwidth}{!}{%
\begin{tabular}{@{}l|l|c|ccc|cc|cc|c@{}}
\toprule
\multirow{2}{*}{Model} & \multirow{2}{*}{LLM} & Frame & \multicolumn{3}{c|}{Efficiency} & \multicolumn{2}{c|}{Params (B)} & \multirow{2}{*}{\shortstack{COIN\\Avg.$\uparrow$}} \\
& & Strategy & TFLOPs & Latency & Dec/s & Total & Active & \\
\midrule
VideoLLM-online-7B-v1~\cite{joya2024videollmonline} & Llama-2-7B & 1 & 9.8 & 2.3 & 28.6 & 6.7 & 6.7 & 60.2 \\
VideoLLM-online-8B-v1+~\cite{joya2024videollmonline} & Llama-3-8B & 1+3x3 & 98.7 & 2.8 & 27.8 & 8.0 & 8.0 & 61.8 \\
VideoLLM-MoD~\cite{enhong2024videollmmod} & Llama-3-8B & 1+3x3 & 60.0 & 3.0 & 24.6 & 8.0 & 8.0 & 61.8 \\
VideoLLM-online-1B-v1+ & Llama-3.2-1B & 1+3x3 & 15.2 & 1.3 & 54.7 & 1.2 & 1.2 & 59.3 \\
\midrule
\rowcolor{oursrow}\textbf{\motemodel{5} (ours)} & Llama-3.2-1B & 1+3x3 & 28.9 & 1.8 & 37.7 & 5.3 & 2.0 & \textbf{62.9} \\
\bottomrule
\end{tabular}
}
\caption{
Efficiency, LLM-only parameter summary and COIN average accuracy. Profiling
uses one NVIDIA A100 with 600 frames, 64 prompt tokens, and 64 generated tokens. Parameter counts include token embeddings and transformer decoder layers, but exclude visual encoders and MLP projectors. \motemodel{5} achieves the highest COIN average while activating only 2B LLM parameters per sample.
}
\label{tab:model_scale_accuracy}
\end{table*}

\paragraph{Prompt-conditioned route selection.} 
To evaluate inference without a supplied task index, we generate 50 prompt variants per COIN task and split them 70/10/20 for training, validation, and testing. A 22.7M-parameter MiniLM-L6-v2 selector reaches 100\% test F1 for each of the five predefined intents, so the resulting end-to-end COIN scores equal those obtained with explicit routes. Selection takes 4.35\,ms per prompt and each cached expert-description embedding occupies 1.5\,KiB. Supplementary Section~\ref{sec:supp_prompt_routing} gives protocol details and examples; ambiguous or compositional prompts remain outside this evaluation.

\subsection{Comparison with Prior COIN Baselines}
\label{sec:comparison_public_baselines}

Table~\ref{tab:coin_public_baselines} compares \motemodel{5} against prior COIN baselines under the same five-task protocol. The upper block contains earlier non-LLM procedural-video methods, while the lower block contains LLM-based VideoLLM variants. \motemodel{5} achieves the highest accuracy on all five COIN tasks, improving the five-task average from 61.8\% for the strongest 8B VideoLLM baseline to 62.9\%.

The public 7B--8B baselines differ from our model in decoder scale, visual encoder, and training recipe. We therefore treat them as protocol-level comparisons rather than isolated architecture controls. Under this conservative interpretation, \motemodel{5} remains competitive: despite using a smaller active LLM path, it exceeds the reported average accuracy of the larger VideoLLM baselines.

To reduce the scale confound, we also include VideoLLM-online-1B-v1+ and VideoLLM-online-3B-v1+, VideoLLM-online baselines trained from the same Llama-3.2 and SigLIP2 family initialization as our default model. In this comparison, the \mote{} conversion achieves the highest five-task average of 62.9\%. The controlled evidence for the \mote{} decoder design therefore comes from two comparisons: the matched 1B/3B VideoLLM-online baselines in Table~\ref{tab:coin_public_baselines}, and the fixed-topology routing controls in Table~\ref{tab:coin_routing_controls}.

Table~\ref{tab:model_scale_accuracy} makes the active-computation advantage explicit. The 8B VideoLLM baselines activate the full 8B-parameter LLM for each sample, whereas \motemodel{5} activates only a 2B LLM path through the selected task expert and shared expert. Relative to VideoLLM-online-8B-v1+, \mote{} reduces the fixed-request cost from 98.7 to 28.9 TFLOPs and latency from 2.8 to 1.8\,s, while increasing decoding throughput from 27.8 to 37.7 tokens/s. It remains slower than the dense 1B baseline because it runs both shared and routed FFNs, in exchange for higher COIN accuracy.

\subsection{Modular Task Adaptation via Expert Addition and Removal}
\label{sec:expert_addition_results}

Table~\ref{tab:expert_addition} evaluates modular task adaptation through sequential expert addition and removal. In the addition phase, we first train the \textit{Step} route and then introduce the \textit{Next}, \textit{Proc.}, \textit{Task}, and \textit{Proc.+} routes one at a time. At each expansion stage, only the newly added task route is trained, while previously trained routes are kept fixed. Consequently, earlier task accuracies remain unchanged across later stages, showing that adding a new task does not overwrite existing task-specific modules. The final addition stage reaches 61.0\% average accuracy over the five active tasks, compared with 62.9\% for the jointly trained \motemodel{5}. This gap reflects a stability--plasticity trade-off: frozen expert addition prioritizes preservation of existing task behavior, whereas joint training allows shared components and task experts to adapt together for higher peak multi-task accuracy. In the removal phase, dropping an expert disables the corresponding task route, while retained routes keep their parameters and measured accuracies. This supports expert removal as a practical deployment mechanism for disabling task-specific capabilities.

\subsection{Cross-Dataset Task Extension}
\label{sec:ego4d_coin_results}

Table~\ref{tab:ego4d_coin} evaluates cross-dataset task extension in a single model. We start from an Ego4D narration model, which contains an online route for deciding when to speak and what narration to generate. We then expand the same model with five COIN experts and train only the newly added COIN routes. The shared video-language backbone, including the visual connector and decoder attention layers, is not retrained on COIN. Thus, the COIN experts operate on shared representations learned from Ego4D narration training, while only the task-specific decoder FFN routes are adapted to the new COIN objectives. This protocol tests whether a model trained for one dataset-specific task family can acquire a new set of procedural tasks while retaining the original online narration capability in the same checkpoint.

The split baseline rows clarify the evaluation scope. VideoLLM-online and VideoLLM-MoD report strong COIN and Ego4D results, but these scores are obtained from separate task-specific finetuned models rather than from one model that supports both task families. By contrast, \motemodel{6} reports COIN and Ego4D metrics from a single expanded checkpoint that retains the Ego4D narration expert while adding five COIN experts. The expanded model remains competitive on the newly added COIN tasks and preserves online narration behavior, achieving the highest Fluency value in the table. The Ego4D-only baselines still obtain lower PPL and TimeDiff, indicating better standalone narration likelihood and timing. This result suggests that expert expansion can reuse a frozen shared backbone when the initial training domain and the added tasks have sufficient semantic overlap. Therefore, the comparison should be interpreted as evidence for backward-compatible task expansion across related video-language datasets, not as a direct claim of superior Ego4D-only narration quality or unrestricted transfer to unrelated domains.

\begin{table*}[t]
\centering
\small
\setlength{\tabcolsep}{3pt}
\resizebox{0.8\textwidth}{!}{%
\begin{tabular}{l|c|cccc|cccc}
\toprule
COIN & Stage 1 & \multicolumn{4}{c|}{Expert addition stages} & \multicolumn{4}{c}{Expert removal stages} \\
Benchmark & Step & \addstage{Next} & \addstage{Proc.} & \addstage{Task} & \addstage{Proc.+} & \remstage{Proc.+} & \remstage{Task} & \remstage{Proc.} & \remstage{Next} \\
\midrule
Step & 63.1 & 63.1 & 63.1 & 63.1 & 63.1 & 63.1 & 63.1 & 63.1 & 63.1 \\
Next & -- & 46.2 & 46.2 & 46.2 & 46.2 & 46.2 & 46.2 & 46.2 & -- \\
Proc. & -- & -- & 48.4 & 48.4 & 48.4 & 48.4 & 48.4 & -- & -- \\
Task & -- & -- & -- & 92.8 & 92.8 & 92.8 & -- & -- & -- \\
Proc.+ & -- & -- & -- & -- & 54.6 & -- & -- & -- & -- \\
\bottomrule
\end{tabular}
}
\caption{
\textbf{Modular task adaptation through sequential expert addition and removal}. We first train a single \textit{Step} expert, then add one task expert at a time for \textit{Next}, \textit{Proc.}, \textit{Task}, and \textit{Proc.+}. After all five experts are available, we remove them in reverse order. Accuracy is reported after each addition or removal stage, and a dash indicates that the corresponding task expert is not present. Stable scores for previously added tasks indicate that adding or removing other experts does not alter their routed inference paths.
}

\label{tab:expert_addition}
\end{table*}

\begin{table*}[t]
\centering
\small
\setlength{\tabcolsep}{3pt}
\resizebox{\textwidth}{!}{%
\begin{tabular}{@{}ll|ccccc|ccc@{}}
\toprule
\multirow{2}{*}{Method} & \multirow{2}{*}{Model instance} & \multicolumn{5}{c|}{COIN Benchmark Top-1 Accuracy$\uparrow$} & \multicolumn{3}{c}{Ego4D Narration Stream Validation} \\
& & Step & Next & Proc. & Task & Proc.+ & PPL$\downarrow$ & TimeDiff$\downarrow$ & Fluency$\uparrow$ \\
\midrule
\multicolumn{10}{@{}l}{\textit{Separate finetuned models reported by prior work}} \\
VideoLLM-online-7B-v1~\cite{joya2024videollmonline} & COIN-only & 59.8 & 48.1 & 47.9 & 92.1 & 52.9 & -- & -- & -- \\
VideoLLM-online-7B-v1~\cite{joya2024videollmonline} & Ego4D-only & -- & -- & -- & -- & -- & 2.43 & 2.25 & 42.6 \\
VideoLLM-online-8B-v1+~\cite{joya2024videollmonline} & COIN-only & 63.1 & 49.1 & 49.8 & 92.7 & 54.1 & -- & -- & -- \\
VideoLLM-online-8B-v1+~\cite{joya2024videollmonline} & Ego4D-only & -- & -- & -- & -- & -- & 2.40 & 2.05 & 45.3 \\
VideoLLM-MoD~\cite{enhong2024videollmmod} & COIN-only & 63.4 & \textbf{49.7} & 49.8 & 92.8 & 53.3 & -- & -- & -- \\
VideoLLM-MoD~\cite{enhong2024videollmmod} & Ego4D-only & -- & -- & -- & -- & -- & \textbf{2.41} & \textbf{2.04} & 45.2 \\
\midrule
\multicolumn{10}{@{}l}{\textit{\motemodel{6} expert expansion in a single model}} \\
\rowcolor{oursrow}\textbf{\motemodel{6} (ours)} & Single expanded model & \textbf{64.9} & 47.6 & \textbf{50.5} & \textbf{93.9} & \textbf{56.4} & 2.57 & 2.25 & \textbf{51.0} \\
\bottomrule
\end{tabular}
}
\caption{
\textbf{Cross-dataset task extension from online narration to offline procedural tasks.} Prior VideoLLM-online and VideoLLM-MoD results are reported from separate COIN-only and Ego4D-only finetuned model instances. In contrast, the final row evaluates one expanded \motemodel{6} checkpoint that starts from an Ego4D narration route and adds five COIN task routes. Ego4D validation reports PPL, TimeDiff, and Fluency.
}
\label{tab:ego4d_coin}
\end{table*}

\subsection{Auxiliary Cross-Domain Evaluation}
\label{sec:aux_cross_domain_eval}
\label{sec:cross_domain_kie}

While our primary experiments focus on procedural video, we include an auxiliary evaluation on Key Information Extraction (KIE) from receipt images to verify if the MoTE FFN conversion mechanism applies outside video tokens. Unlike video benchmarks, this task requires parsing text-rich, static documents into structured outputs. We adapted the dense GLM-OCR-0.9B \cite{glmocr2026} backbone into the MoTE framework by introducing a shared expert alongside two task-specific routes: a retained Optical Character Recognition (OCR) route for the original capabilities and a newly trained KIE route. The KIE expert was trained on the SROIE and CORD datasets to generate structured field predictions directly from raw document images without an intermediate text-extraction pipeline.

The converted \glmocrmote{} model achieves 87.90\% Micro-F1 on SROIE KIE and 95.79\% Micro-F1 on CORD, successfully raising the dense GLM-OCR CORD and SROIE baseline from 20.07\% and 55.04\% respectively to demonstrate effective adaptation to structured extraction. Crucially, the retained OCR route perfectly preserves the original model's performance on OmniDocBench v1.5 after conversion and KIE training. Full per-field metrics are provided in Supplementary Tables~\ref{tab:supp_sroie_and_ocr_parity_grouped} and~\ref{tab:supp_cord_clean_comparison}.

By leveraging the MoTE architecture, the resulting \glmocrmote{} proves competitive with several open multimodal models, demonstrating that task-specific experts can successfully attach to non-video backbones to introduce entirely new capabilities without inducing catastrophic forgetting. We present these findings primarily as auxiliary evidence of the architecture's cross-domain applicability.

\begin{table*}[t]
\centering
\small
\setlength{\tabcolsep}{4pt}
\begin{minipage}[t]{0.48\textwidth}
\centering
\resizebox{\linewidth}{!}{%
\begin{tabular}{@{}l|c|c@{}}
\toprule
Method & Input Modality & Micro-F1 (\%) $\uparrow$ \\
\midrule
LayoutLLM-7B~\cite{chu2023layoutllm} & Image+Text+Layout & 72.12 \\
DocKylin~\cite{zhang2024dockylin} & Image + Text & 69.60 \\
Qwen-VL-7B~\cite{bai2023qwenvl} & Image + Text & 58.59 \\
LLaVA-1.5-7B~\cite{liu2023llava15} & Image + Text & 3.83 \\
LLaVAR-7B~\cite{zhang2023llavar} & Image + Text & 2.38 \\
\midrule
GLM-OCR-0.9B~\cite{glmocr2026} & Image & 55.04 \\
\midrule
\rowcolor{oursrow}\textbf{\glmocrmote{} (ours)} & \textbf{Image} & \textbf{87.90} \\
\bottomrule
\end{tabular}
}
\caption{SROIE KIE Micro-F1 comparison for image-using models. \glmocrmote{} reaches 87.90\%, compared with 72.12\% for the strongest image-using baseline in this compact table.}
\label{tab:sroie_and_ocr_parity_grouped}
\end{minipage}
\hfill
\begin{minipage}[t]{0.48\textwidth}
\centering
\resizebox{\linewidth}{!}{%
\begin{tabular}{@{}l|c|c@{}}
\toprule
Method & Input Modality & Micro-F1 (\%) $\uparrow$ \\
\midrule
UDOP~\cite{tang2023udop} & Image+Text+Layout & 97.60 \\
LayoutLMv2~\cite{xu2021layoutlmv2} & Image+Text+Layout & 96.01 \\
DocTr~\cite{liao2023doctr} & Image+Text+Layout & 94.40 \\
Donut~\cite{kim2022donut} & Image + Text & 91.60 \\
\midrule
GLM-OCR-0.9B~\cite{glmocr2026} & Image & 20.07 \\
\midrule
\rowcolor{oursrow}\textbf{\glmocrmote{} (ours)} & \textbf{Image} & \textbf{95.79} \\
\bottomrule
\end{tabular}
}
\caption{CORD KIE Micro-F1 comparison for image-using models. \glmocrmote{} raises the GLM-OCR baseline from 20.07\% to 95.79\%, approaching the strongest multimodal baseline while retaining the OCR route discussed in the text.}
\label{tab:cord_clean_comparison}
\end{minipage}
\end{table*}

\subsection{Ablation Study}
\label{sec:controlled_ablations}

\subsubsection{Effect of Routing Granularity}
\label{sec:comparison_routing_controls}

Table~\ref{tab:coin_routing_controls} compares how the same five-expert topology behaves under different routing choices. The dense control removes sparsity by activating every routed expert for every sample. The token-level control learns an independent top-1 expert for each token. The sample-level control learns one top-1 expert for the full video--prompt sample, matching \mote{} at the sample granularity but without explicit task supervision. \mote{} instead uses the known task identity to select the task-aligned expert. The experiment therefore tests whether task-aligned sample routing is more effective than dense expert fusion or learned sparse routing at token and sample granularity.

\begin{table}[t]
\centering
\scriptsize
\setlength{\tabcolsep}{3pt}
\resizebox{0.9\columnwidth}{!}{%
\begin{tabular}{l|ccccc|c}
\toprule
\multirow{2}{*}{Method} & \multicolumn{5}{c|}{COIN Benchmark Top-1 Accuracy$\uparrow$} & \multirow{2}{*}{Avg.} \\
& Step & Next & Proc. & Task & Proc.+ & \\
\midrule
VideoLLM-1B+5E-Dense & 64.4\,\dzero & 49.8\,\dzero & 48.7\,\dzero & 93.2\,\dzero & 51.5\,\dzero & 61.5\,\dzero \\
VideoLLM-MoE-Top1-token & 63.7\,\dneg{0.7} & 49.5\,\dneg{0.3} & 48.5\,\dneg{0.2} & 93.8\,\dpos{0.6} & 51.7\,\dpos{0.2} & 61.4\,\dneg{0.1} \\
VideoLLM-MoE-Top1-sample & 64.9\,\dpos{0.5} & 50.3\,\dpos{0.5} & 49.2\,\dpos{0.5} & 93.6\,\dpos{0.4} & 52.3\,\dpos{0.8} & 62.1\,\dpos{0.6} \\
\rowcolor{oursrow}\textbf{\motemodel{5} (ours)} & \textbf{65.1\,\dpos{0.7}} & \textbf{50.3\,\dpos{0.5}} & \textbf{50.5\,\dpos{1.8}} & \textbf{94.4\,\dpos{1.2}} & \textbf{54.2\,\dpos{2.7}} & \textbf{62.9\,\dpos{1.4}} \\
\bottomrule
\end{tabular}
}
\caption{
\textbf{Effect of Routing Granularity}. Numbers in parentheses are accuracy changes relative to the VideoLLM-1B+5E-Dense control; green and red indicate positive and negative changes, respectively. VideoLLM-MoE-Top1-token learns token-level routes, VideoLLM-MoE-Top1-sample learns one route for the full sample, and \motemodel{5} uses explicit task-index routing.
}
\label{tab:coin_routing_controls}
\end{table}

With the same converted decoder capacity, \motemodel{5} achieves 62.9\% average accuracy, compared with 61.5\% accuracy of VideoLLM-1B+5E-Dense for dense all-expert activation and 61.4\% for token routing. This indicates that task structure is useful for choosing the FFN transformation: activating all routed experts dilutes specialization, while token-level routing breaks the task boundary across the generated sequence. The advantage is especially clear on procedure-oriented tasks, where the requested output depends on a longer temporal horizon and a more structured response format.

The sample-level baseline is closest to \mote{} in routing granularity because both choose one expert route for the whole video--prompt sample. The difference is route semantics: sample-level MoE learns a latent top-1 route, whereas \mote{} routes by task identity. Sample-level routing improves over dense activation, but \mote{} gives the strongest average accuracy and the largest gains on structured procedure prediction. This suggests that, for task-structured procedural video understanding, MoTE is most effective when the route is aligned with the user-facing task.

\subsubsection{Effect of Expert Construction Method}
\label{sec:expert_construction_results}

We compare the two expert construction methods in Table~\ref{tab:expert_construction}. Instead of copying the full pretrained FFN into each expert, \textbf{VideoLLM-MoTE-8B+5E-Split} uses the same 8B VideoLLM backbone and partitions each FFN into six equally sized experts, corresponding to one shared expert and five task experts, following LLaMA-MoE~\cite{zhu2024llamamoe}. This keeps the experts disjoint inside each FFN rather than giving each task route a full copied FFN.

VideoLLM-MoTE-8B+5E-Split reaches 60.3\% average accuracy, below the 62.9\% average of the full-FFN-copy \motemodel{5}. This suggests that task-routed video-language experts benefit from retaining the full pretrained FFN transformation in each route before specialization. Splitting neurons into small disjoint task groups is parameter efficient, but it reduces the per-task FFN capacity available for procedural output formats and long-horizon prediction.

\begin{table}[t]
\centering
\small
\setlength{\tabcolsep}{4pt}
\resizebox{0.9\columnwidth}{!}{%
\begin{tabular}{l|c|ccccc|c}
\toprule
\multirow{2}{*}{Method} & \multirow{2}{*}{MoE Construction} & \multicolumn{5}{c|}{COIN Benchmark Top-1 Accuracy$\uparrow$} & \multirow{2}{*}{Avg.} \\
&& Step & Next & Proc. & Task & Proc.+ & \\
\midrule
VideoLLM-MoTE-8B+5E-Split & Splitting FFN & 63.3 & 48.8 & 48.3 & 92.9 & 48.3 & 60.3 \\
\rowcolor{oursrow}\motemodel{5} (default) & Copying FFN & \textbf{65.1} & \textbf{50.3} & \textbf{50.5} & \textbf{94.4} & \textbf{54.2} & \textbf{62.9} \\
\bottomrule
\end{tabular}
}
\caption{
\textbf{Effect of Expert Construction Method}. VideoLLM-MoTE-8B-5E-Split partitions each VideoLLM FFN into six equally sized expert groups (one
shared expert and five task experts), while \motemodel{5} initializes each shared and task expert as a full copied FFN. Copying FFN layer into multiple experts achieves better performance.
}
\label{tab:expert_construction}
\end{table}

\subsection{Discussion}
\label{sec:experiments_summary}

The experiments support a simple claim: task identity is a useful structure for multi-task settings. Under the same expert topology, explicit task-index routing gives the strongest average accuracy across our tested benchmarks among the internal controls and uses the expert capacity more effectively than dense all-expert activation or learned one-expert sparse routing. The clearest gains appear on tasks with highly specialized objectives, where a single shared transformation is more likely to mix incompatible task requirements. This suggests that task routing helps most when task boundaries align with genuinely different computational needs.

The results also clarify where specialization occurs. In \mote, attention and foundational representation alignment remain shared, while task-specific transformations are assigned to the feed-forward experts. This separates shared context modeling from task-dependent computation. This design sits between full parameter sharing and full model duplication: per-sample active expert computation is constant, but stored parameters grow linearly as task experts are added.

Finally, the expert-addition experiments expose a stability-plasticity tradeoff: freezing old routes preserves their measured behavior but gives lower peak accuracy than joint training. The auxiliary GLM-OCR-0.9B experiment provides a second, non-video test of the MoTE conversion. Adding a receipt-KIE route raises CORD micro-F1 from 20.07\% to 95.79\% from raw images while the retained route matches the reported OmniDocBench OCR metrics. Together, these results support backward-compatible modular expansion in the tested video and document settings.

\section{Conclusion}
\label{sec:conclusion}

We introduced \mote~(Mixture of Task Experts), a modular approach for extending LLM decoders by transforming their FFNs into task-specific experts while preserving a shared backbone. This design isolates task-specialized knowledge, activates only the expert required for a given task, and enables new capabilities to be added without modifying existing expert routes. Across five COIN video-understanding tasks, VideoLLM-MoTE consistently outperforms dense all-expert activation, learned sparse-routing variants, and reported larger VideoLLM baselines, despite using fewer active parameters. Our auxiliary document-understanding experiment further demonstrates the value of this modularity: MoTE can acquire new structured-output capabilities while fully retaining its foundational OCR performance. Together, these results show that explicit task specialization and route preservation provide an effective and parameter-efficient alternative to monolithic model expansion in the evaluated settings.

\section*{Acknowledgments}
This work has been partially funded by the European Union, under Horizon Europe, grant number 101092889, project VICTOR-XR.

\bibliography{video_literature}

\clearpage
\appendix

\section{Supplementary Material}
\label{sec:supplementary}

The supplementary material follows the experimental workflow. Sections~\ref{sec:supp_architecture_params}--\ref{sec:supp_bucketed_sampling} document model accounting, training configuration, and data sampling. Section~\ref{sec:supp_eval_scheme} defines the evaluation protocols. Section~\ref{sec:supp_prompts} presents the rendered model inputs, and Section~\ref{sec:supp_prompt_routing} groups the expert-description registry with the prompt-conditioned routing evaluation. Section~\ref{sec:supp_expert_swap} then reports the expert-swap diagnostic. Section~\ref{sec:supp_document_eval} presents the auxiliary document architecture, protocol, and full benchmark results. Section~\ref{sec:supp_limitations} concludes with scope limitations.

\subsection{MoTE LLM Parameter Accounting}
\label{sec:supp_architecture_params}

Table~\ref{tab:model_scale_accuracy} reports LLM-only counts, excluding the shared visual encoder and projector. From the existing Llama-3.2-1B checkpoint, the required module totals are
\[
P_{\mathrm{dense}}=1{,}235{,}814{,}400,
\qquad
P_{\mathrm{FFN}}=805{,}306{,}368,
\]
where $P_{\mathrm{FFN}}$ sums the FFNs across all 16 layers. These checkpoint-derived totals are sufficient for the \mote{} calculation.

Let $T$ be the number of stored routed experts, $S$ the number of shared experts, and $k$ the number of routed experts activated for one sample. Because $P_{\mathrm{dense}}$ already contains one FFN path, replacing it with $T+S$ copies gives
\[
\begin{aligned}
P_{\mathrm{total}}(T,S)
&=P_{\mathrm{dense}}+(T+S-1)P_{\mathrm{FFN}},\\
P_{\mathrm{active}}(k,S)
&=P_{\mathrm{dense}}+(k+S-1)P_{\mathrm{FFN}}.
\end{aligned}
\]
For \motemodel{5}, $T=5$, $S=1$, and top-1 routing gives $k=1$:
\[
\begin{aligned}
P_{\mathrm{total}}
&=1{,}235{,}814{,}400+5\cdot805{,}306{,}368\\
&=5{,}262{,}346{,}240 \simeq 5.3\mathrm{B},\\
P_{\mathrm{active}}
&=1{,}235{,}814{,}400+805{,}306{,}368\\
&=2{,}041{,}120{,}768 \simeq 2.0\mathrm{B}.
\end{aligned}
\]
These exact values round to the 5.3B total and 2.0B active counts in Table~\ref{tab:model_scale_accuracy}. Adding stored task routes increases total parameters linearly by $P_{\mathrm{FFN}}$, whereas the active count remains fixed because each sample executes only the shared and one routed FFN path.

\subsection{Training Configuration}
\label{sec:supp_hyperparams}

Table~\ref{tab:supp_hyperparams} lists the main hyperparameters for the \videollmmote{} experiments. For COIN, the tasks are trained as offline answer-generation problems: visual-frame positions are masked from the streaming silence loss and supervision covers the assistant answer plus EOS. Ego4D narration keeps the streaming silence term, so the narration route learns to produce short narrations at annotated response points and to remain silent elsewhere in the stream. In the main COIN results, task labels provide the route assignment directly.

\begin{table}[!htbp]
\centering
\footnotesize
\setlength{\tabcolsep}{4pt}
\renewcommand{\arraystretch}{1.06}
\begin{tabular}{@{}>{\raggedright\arraybackslash\ttfamily}p{0.34\linewidth}>{\raggedright\arraybackslash\normalfont}p{0.58\linewidth}@{}}
\toprule
\textbf{Hyper-parameter} & \textbf{Value} \\
\midrule
\texttt{frame\_fps} / \texttt{max\_frames} & 2 / 1200 \\
GPUs & 4$\times$A100-80GB \\
\texttt{per\_device\_batch} & 1 train / 1 eval \\
\texttt{gradient\_acc\_steps} & 8 \\
Effective train batch & 32 samples \\
\texttt{num\_train\_epochs} & 10 \\
\texttt{learning\_rate} & 1e-4 \\
\texttt{optimizer} & AdamW \\
\texttt{lr\_scheduler\_type} & cosine \\
\texttt{warmup\_ratio} & 0.05 \\
\texttt{lora\_targets} & q/k/v/o, gate/up/down, lm\_head \\
\texttt{finetune\_modules} & MLP projector \\
\texttt{lora\_r} / \texttt{lora\_alpha} & 128 / 256 \\
\texttt{lora\_dropout} & 0.05 \\
\texttt{moe\_router\_aux\_loss} & 0.0 \\
\texttt{attn\_implementation} & sdpa \\
\bottomrule
\end{tabular}
\caption{Hyper-parameters used for training \videollmmote.}
\label{tab:supp_hyperparams}
\end{table}

\subsection{Data Sampling Strategy}
\label{sec:supp_bucketed_sampling}

The routed \mote{} decoder activates a task expert according to the expert id attached to each training sample. The expert buckets are not equally sized: for example, step recognition, next-step forecasting, task recognition, and procedure forecasting can contribute different numbers of training examples. A uniform-over-experts schedule would repeatedly recycle smaller buckets and overtrain their experts, while larger buckets would be under-sampled relative to the evidence available for those tasks. Conversely, unconstrained mixed shuffling can give different data-parallel ranks different routed experts at the same local step.

\begin{figure}[!htbp]
    \centering
    \includegraphics[
    width=0.94\textwidth,
    trim={0cm 0cm 0cm 2.3cm},
    clip
    ]{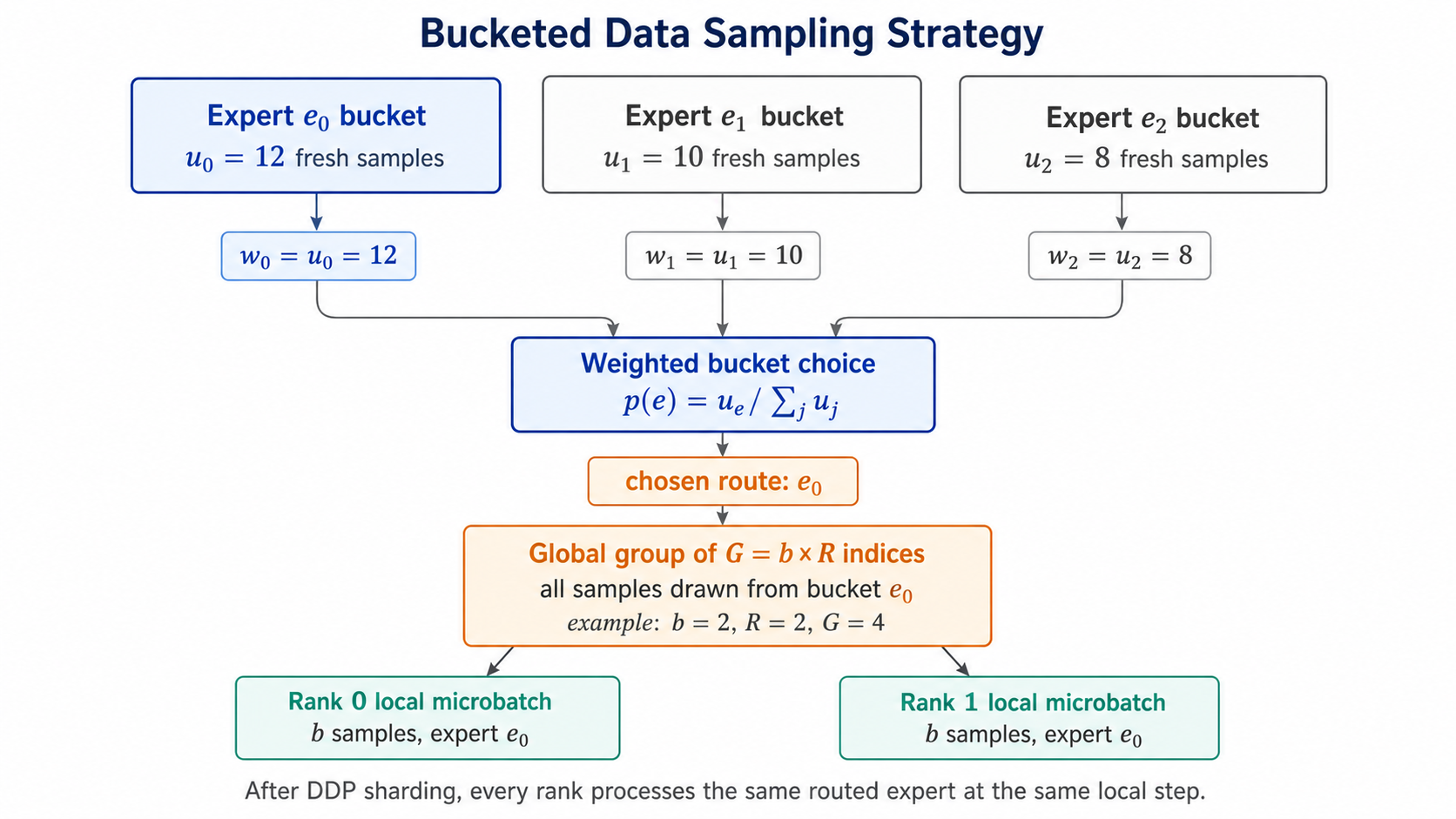}
    \caption{\textbf{Bucketed rank-aligned sampling.} The sampler groups a concatenated multi-task dataset into expert buckets. At each local training step, it samples one expert bucket with probability proportional to the number of fresh samples remaining in that bucket, draws one global group of $G=bR$ indices from the selected bucket, and lets distributed sharding split the group into rank-local microbatches. Here $b$ is the local batch size and $R$ is the number of data-parallel ranks.}
    \label{fig:supp_bucketed_sampling}
\end{figure}

Figure~\ref{fig:supp_bucketed_sampling} illustrates the bucketed rank-aligned strategy used in our training runs. At the start of an epoch, samples are grouped by their expert id and each bucket is shuffled. At each local step, the sampler chooses one expert bucket with probability proportional to the number of fresh samples still unseen in that bucket. It then draws a global group of $G=bR$ indices from that selected bucket, where $b$ is the per-rank local batch size and $R$ is the number of data-parallel ranks. Distributed sharding splits this contiguous group across ranks, so every rank receives a local microbatch for the same routed expert at that step. When a selected bucket is exhausted, it is reshuffled and reused only as needed. This preserves the task-size bias of the training set while reducing both overexposure of small buckets and underexposure of large buckets.

\subsection{Evaluation Protocols}
\label{sec:supp_eval_scheme}

\paragraph{COIN benchmarks.}
COIN outputs are decoded as free-form text and then mapped back to the task label space before scoring. For Step, Next, and Task, we lowercase the response, remove a trailing period, and first check for an exact match with the target label. If no exact match is found, the response is assigned to the closest valid class in the task taxonomy using Levenshtein distance. Top-1 accuracy is then computed over the mapped labels. For Proc. and Proc.+, the generated numbered list is split into individual lines, optional number prefixes are removed, and each predicted future step is mapped to the nearest COIN step label before computing step-wise top-1 accuracy over aligned future-action positions.

\paragraph{Ego4D narration stream.}
Ego4D narration is evaluated as an online stream rather than as an offline classification problem. Each narration turn has an annotated response time and a short target narration. We report three diagnostics following the prior online VideoLLM protocol: PPL measures token-level narration likelihood, TimeDiff measures the absolute temporal offset between the predicted and annotated response time in seconds, and Fluency combines timing and token correctness into a percentage score. Lower PPL and TimeDiff are better, while higher Fluency is better. 

\paragraph{Receipt KIE and OCR preservation.}
The auxiliary document experiment is evaluated separately from the video benchmarks. For SROIE Task 3, we report strict field-level F1 for company, date, address, and total, together with micro-F1 over the receipt fields in Table~\ref{tab:supp_sroie_and_ocr_parity_grouped}{}. For CORD, we report relaxed field-level F1 over the receipt categories used in the benchmark table and the corresponding micro-F1 in Table~\ref{tab:supp_cord_clean_comparison}. The OCR route is checked on OmniDocBench v1.5 using the original GLM-OCR metrics: overall score, text edit distance, formula score, table score, table-structure score, and reading-order edit distance. These document metrics are used to test route adaptation and route preservation; they are not averaged with the video-language results.

\subsection{Prompt Templates}
\label{sec:supp_prompts}

We show examples of our prompts to illustrate the dialogue format. \textcolor{promptblue}{\vtok} markers are placeholders for visual tokens and orange text marks supervised assistant response during training.

The concrete targets below use official annotation labels from a \texttt{\textcolor{promptblue}{Make french fries}} video. For Ego4D narration, the stream remains online: an initial frame chunk is followed by the narration instruction, then additional frame chunks and timestamped assistant narrations.

\promptcard{System Prompt}{applies to all routes}{%
\chatdump{\textcolor{promptlabel}{A multimodal AI assistant is helping users with some activities. Below is their conversation, interleaved with the list of video frames received by the assistant.}}
}

\promptcard{COIN Step Recognition}{Step route}{%
\chatdump{\chatbos{} \systempromptref\par\par
User: What is the action in the video? Format your answer concisely. No extra text output.\par\par
\textcolor{promptblue}{[\vtok\ldots\vtok,\vtok\ldots\vtok,\ldots]}\par
Assistant: \assistantspan{Cut potato into strips.}\chateot}
}

\promptcard{COIN Next-Step Forecasting}{Next route}{%
\chatdump{\chatbos{} \systempromptref\par\par
User: What is the next action for the video? Format your answer concisely. No extra text output.\par\par
\textcolor{promptblue}{[\vtok\ldots\vtok,\vtok\ldots\vtok,\ldots]}\par
Assistant: \assistantspan{Soak them in water.}\chateot}
}

\promptcard{COIN Task Recognition}{Task route}{%
\chatdump{\chatbos{} \systempromptref\par\par
User: What is the overall activity in the video? Format your answer concisely. No extra text output.\par\par
\textcolor{promptblue}{[\vtok\ldots\vtok,\vtok\ldots\vtok,\ldots]}\par
Assistant: \assistantspan{Make french fries.}\chateot}
}

\promptcard{COIN Procedure Forecasting}{Proc. route}{%
\chatdump{\chatbos{} \systempromptref\par\par
User: What is the next 3 actions for the video? Format your answer concisely, listing each action on a new line with a number prefix. No extra text output.\par\par
\textcolor{promptblue}{[\vtok\ldots\vtok,\vtok\ldots\vtok,\ldots]}\par
Assistant: \assistantspan{1. Soak them in water.}\par
\assistantspan{2. Dry strips.}\par
\assistantspan{3. Put in the oil to fry.}\chateot}
}

\promptcard{COIN Procedure Forecasting with Task Goal}{Proc.+ route}{%
\chatdump{\chatbos{} \systempromptref\par\par
User: To \texttt{\textcolor{promptblue}{Make french fries}}, what is the next 3 actions for the video? Format your answer concisely, listing each action on a new line with a number prefix. No extra text output.\par\par
\textcolor{promptblue}{[\vtok\ldots\vtok,\vtok\ldots\vtok,\ldots]}\par
Assistant: \assistantspan{1. Soak them in water.}\par
\assistantspan{2. Dry strips.}\par
\assistantspan{3. Put in the oil to fry.}\chateot}
}

\promptcard{Ego4D Narration Stream}{Narration route}{%
\chatdump{\chatbos{} \systempromptref\par\par
\textcolor{promptblue}{[\vtok\ldots\vtok]}\par
User: Please concisely narrate the video in real time. Use the tag 'C' to denote the camera wearer, and other letter tags, such as 'X', to denote other individuals in the scene.\par
\textcolor{promptblue}{[\vtok\ldots\vtok,\vtok\ldots\vtok]}\par
Assistant: \assistantspan{C walks around a room.}\chateot\par
\textcolor{promptblue}{[\vtok\ldots\vtok,\vtok\ldots\vtok,\vtok\ldots\vtok]}\par
Assistant: \assistantspan{C picks up a wire from the floor.}\chateot}
}

\subsection{Prompt-conditioned Routing Evaluation}
\label{sec:supp_prompt_routing}

Table~\ref{tab:supp_expert_descriptions} records the natural-language registry used by the prompt-conditioned selector. These descriptions are not prompts shown to the decoder; they define the expert-selection component described in Section~\ref{sec:method_routing}.

\begin{table}[!htbp]
\centering
\footnotesize
\setlength{\tabcolsep}{4pt}
\renewcommand{\arraystretch}{1.06}
\begin{tabular}{@{}p{0.2\linewidth}p{0.76\linewidth}@{}}
\toprule
\textbf{Task} & \textbf{Expert natural-language description} \\
\midrule
Current Step & Recognize the action currently being performed in the observed video segment; use this route for prompts asking what is happening now or which procedural step is visible. \\
Next Step & Predict the immediate next action that should follow the observed segment; use this route for short-horizon forecasting prompts asking what comes next. \\
Task & Identify the overall instructional goal or activity category of the video, independent of the current substep; use this route for prompts asking for the complete task being performed. \\
Procedure & Forecast a sequence of upcoming procedural actions from the current visual state without being given an explicit task goal; use this route for open-ended procedure-continuation prompts. \\
Task Procedure & Forecast upcoming procedural actions while conditioning on a provided goal or task name; use this route when the prompt states the target activity and asks for future steps. \\
Narration & Generate concise timestamp-aligned egocentric narrations for streaming video; use this route for online narration prompts that require deciding both what to say and when to say it. \\
\bottomrule
\end{tabular}
\caption{Task-to-description mapping used for prompt-conditioned expert selection.}
\label{tab:supp_expert_descriptions}
\end{table}

Because no real user-prompt--task dataset is available for the five COIN intents, GPT-5.5 was used to generate 50 paraphrased prompts per task, giving 250 prompts in total. Each task is split 70/10/20 into 35 training, 5 validation, and 10 held-out test prompts. We instantiate the selector with MiniLM-L6-v2 (22.7M parameters) and evaluate the predicted top-1 task route. It reaches 100\% test F1 for each task; consequently, end-to-end COIN scores under prompt-selected routing equal those reported with explicit task indices. Selection takes 4.35\,ms per prompt, and each cached expert-description embedding occupies 1.5\,KiB.

The held-out confusion matrix below makes the selector result explicit. All 10 test prompts for each intent are assigned to the correct route, yielding 100\% precision, recall, and F1 for every class (50/50 correct overall).

\begin{center}
\begin{minipage}{0.82\linewidth}
\centering
\footnotesize
\textbf{Held-out prompt-routing confusion matrix (counts)}\par
\vspace{0.35em}
\setlength{\tabcolsep}{5pt}
\renewcommand{\arraystretch}{1.05}
\begin{tabular}{l|ccccc|c}
\toprule
\multirow{2}{*}{Actual route} & \multicolumn{5}{c|}{Predicted route} & \multirow{2}{*}{F1 (\%)} \\
& Step & Next & Proc. & Task & Proc.+ & \\
\midrule
Step  & \cellcolor{oursrow}10 & 0 & 0 & 0 & 0 & 100 \\
Next  & 0 & \cellcolor{oursrow}10 & 0 & 0 & 0 & 100 \\
Proc. & 0 & 0 & \cellcolor{oursrow}10 & 0 & 0 & 100 \\
Task  & 0 & 0 & 0 & \cellcolor{oursrow}10 & 0 & 100 \\
Proc.+ & 0 & 0 & 0 & 0 & \cellcolor{oursrow}10 & 100 \\
\midrule
Macro average & \multicolumn{5}{c|}{50/50 correct} & 100 \\
\bottomrule
\end{tabular}
\par\vspace{0.25em}
\textit{Rows denote ground-truth intents; columns denote selected expert routes.}
\end{minipage}
\end{center}

Table~\ref{tab:supp_prompt_routing_examples} shows three of the 50 generated variants for each task. The examples use diverse phrasing while preserving clear task intent. This experiment is therefore a controlled selector sanity check rather than evidence for routing ambiguous, overlapping, or compositional requests.

\begin{table}[!htbp]
\centering
\scriptsize
\setlength{\tabcolsep}{7pt}
\renewcommand{\arraystretch}{1.14}
\begin{tabular}{@{}>{\raggedright\arraybackslash\ttfamily}p{0.16\linewidth}>{\raggedright\arraybackslash\normalfont}p{0.77\linewidth}@{}}
\toprule
\normalfont\textbf{Task route} & \textbf{Representative generated prompt variants} \\
\midrule
\rowcolor{promptbg}Step & \textcolor{promptlabel}{\textbf{1.}} Which action is visible in this clip? Use a minimal answer with no extra text.\par
\textcolor{promptlabel}{\textbf{2.}} Recognize the step currently shown in the video. Use a minimal answer with no extra text.\par
\textcolor{promptlabel}{\textbf{3.}} What action does this video segment depict? Answer briefly with only the action name.\par
\textcolor{promptlabel}{\emph{\ldots{} 47 additional variants}} \\
Next & \textcolor{promptlabel}{\textbf{1.}} Name the step that follows the current video. Give just the next action label.\par
\textcolor{promptlabel}{\textbf{2.}} What should the person do next? Keep it short and direct.\par
\textcolor{promptlabel}{\textbf{3.}} What is the upcoming action after the video segment? Keep it short and direct.\par
\textcolor{promptlabel}{\emph{\ldots{} 47 additional variants}} \\
\rowcolor{promptbg}Task & \textcolor{promptlabel}{\textbf{1.}} What larger task does this video demonstrate? Give just the overall task label.\par
\textcolor{promptlabel}{\textbf{2.}} What is the person trying to accomplish overall? Return a short activity phrase only.\par
\textcolor{promptlabel}{\textbf{3.}} State the overall procedure represented by the clip. Answer concisely with only the task name.\par
\textcolor{promptlabel}{\emph{\ldots{} 47 additional variants}} \\
Proc. & \textcolor{promptlabel}{\textbf{1.}} Name the upcoming \texttt{\{num\_steps\}} actions in order. Return only the ordered action list.\par
\textcolor{promptlabel}{\textbf{2.}} What sequence of \texttt{\{num\_steps\}} actions follows this segment? Keep the list concise and include no commentary.\par
\textcolor{promptlabel}{\textbf{3.}} Determine the next \texttt{\{num\_steps\}} steps after the shown action. Keep the list concise and include no commentary.\par
\textcolor{promptlabel}{\emph{\ldots{} 47 additional variants}} \\
\rowcolor{promptbg}Proc.+ & \textcolor{promptlabel}{\textbf{1.}} Using \texttt{\{task\}} as the overall activity, name the upcoming \texttt{\{num\_steps\}} actions. Do not include any extra text beyond the answer.\par
\textcolor{promptlabel}{\textbf{2.}} For \texttt{\{task\}}, identify the next action based on the current video. Return only the action or numbered action list.\par
\textcolor{promptlabel}{\textbf{3.}} In the context of \texttt{\{task\}}, what step comes after the shown segment? Use numbered lines when multiple actions are requested.\par
\textcolor{promptlabel}{\emph{\ldots{} 47 additional variants}} \\
\bottomrule
\end{tabular}
\caption{Representative generated prompt variants used to evaluate prompt-conditioned expert selection. Three of 50 variants are shown per task; the held-out split achieves the 100\% test F1 reported in the text.}
\label{tab:supp_prompt_routing_examples}
\end{table}

\clearpage
\subsection{Offline Expert-Swap Routing}
\label{sec:supp_expert_swap}

Table~\ref{tab:supp_expert_swap} provides an offline diagnostic for whether COIN task behavior is localized in the routed experts. We keep the evaluation prompts and route ids unchanged, but swap the trained FFN weights of two task experts across all decoder layers before evaluation. If the experts behave as task-specialized modules, the degradation should concentrate on the swapped tasks while the untouched routes remain close to the base model.

\begin{table}[!h]
\centering
\small
\setlength{\tabcolsep}{4pt}
\resizebox{\width}{!}{%
\begin{tabular}{l|ccccc}
\toprule
\multirow{2}{*}{Method} & \multicolumn{5}{c}{COIN Benchmark Top-1 Accuracy$\uparrow$} \\
& Step & Next & Proc. & Task & Proc.+ \\
\midrule
\rowcolor{oursrow}\motemodel{5} & 65.1 & 50.3 & 50.5 & 94.4 & 54.2 \\
Swap Step$\leftrightarrow$Next & \swapres{21.5} & \swapres{17.4} & 50.5 & 94.4 & 54.2 \\
Swap Next$\leftrightarrow$Proc. & 65.1 & \swapres{23.3} & \swapres{30.7} & 94.4 & 54.2 \\
Swap Proc.$\leftrightarrow$Task & 65.1 & 50.3 & \swapres{4.97} & \swapres{15.36} & 54.2 \\
Swap Task$\leftrightarrow$Proc.+ & 65.1 & 50.3 & 50.5 & \swapres{15.98} & \swapres{5.36} \\
\bottomrule
\end{tabular}
}
\caption{
Offline expert-swap routing diagnostic on COIN. Highlighted cells indicate the two task columns whose expert weights are swapped at evaluation time. Untouched task routes keep the base route weights.
}
\label{tab:supp_expert_swap}
\end{table}

The diagnostic shows that accuracy collapses primarily on the swapped task pair. Swapping Step and Next damages only those two columns, while Proc., Task, and Proc.+ remain at the base values. The same locality appears for the other pairs, with the strongest failures occurring when Proc. is swapped with Task or when Task is swapped with Proc.+. This pattern supports the interpretation that routed FFN experts capture task-specific transformations rather than acting as interchangeable capacity. The diagnostic is not an expert-removal or unlearning evaluation; it is a stress test of expert specialization under deliberately incorrect route-to-weight assignments.

\clearpage
\subsection{Document-Domain Evaluation}
\label{sec:supp_document_eval}

\subsubsection{Architecture}
\label{sec:supp_document_flow}

\begin{figure}[!htbp]
\centering
\includegraphics[width=0.98\textwidth]{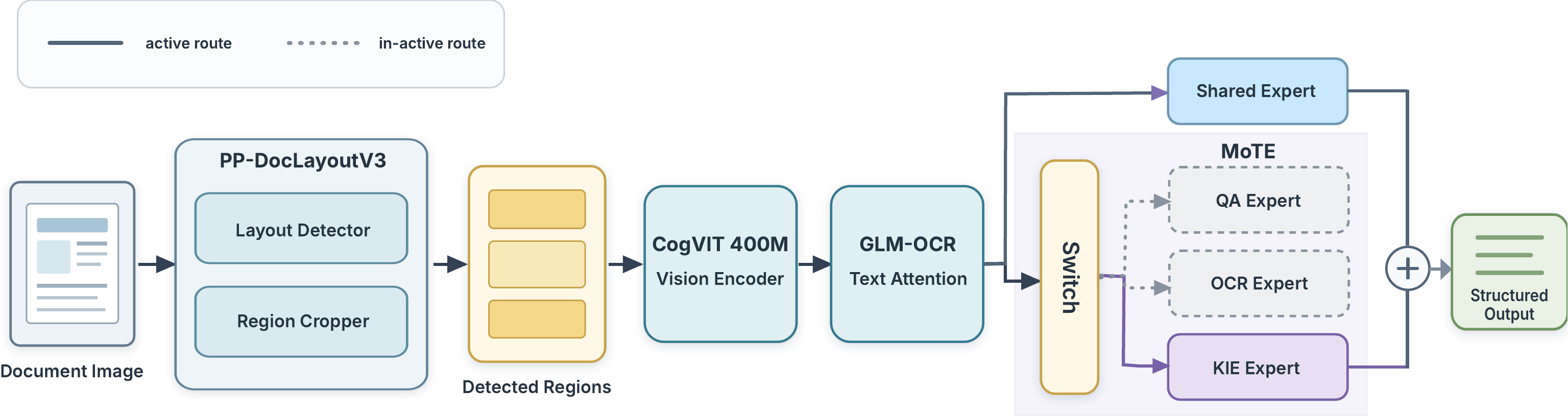}
\caption{\textbf{GLM-OCR-MoTE document processing architecture.} Detected document regions pass through the shared CogVIT encoder and GLM-OCR attention path before the switch activates one task-specific MLP expert alongside the shared expert. Solid lines denote active routes and dotted lines denote inactive routes.}
\label{fig:supp_document_flow}
\label{fig:glm_mote_architecture}
\end{figure}

Figure~\ref{fig:supp_document_flow} expands the auxiliary document experiment from Section~\ref{sec:cross_domain_kie}. PP-DocLayoutV3 detects and crops document regions, CogVIT encodes the resulting visual inputs, and the GLM-OCR attention path forms a shared representation. The \mote{} conversion is confined to the MLP/FFN path: the shared expert is always active, while the switch selects one task expert for the sample. Their outputs are combined  before structured decoding; dotted routes denote experts that are inactive in the illustrated example.

\subsubsection{Detailed Evaluation}
\label{sec:supp_document_kie}

This section expands the compact document-domain tables from Section~\ref{sec:cross_domain_kie}. The \glmocrmote{} conversion uses the same high-level pattern as the video model: an always-active shared FFN path is paired with one selected task route. In this auxiliary setting, the two routed experts correspond to OCR and receipt KIE. During KIE adaptation, the new receipt route learns to emit structured receipt fields directly from image inputs, while the retained OCR route is evaluated separately on the original OCR benchmark.

The comparison groups baselines by input modality because this determines how much document structure is provided to the model. OCR-text-only and OCR-plus-layout baselines receive recognized text, bounding boxes, or both. Image-using baselines receive pixels, sometimes together with OCR or layout. \glmocrmote{} is listed as an image-input model because the KIE route predicts fields from the receipt image without an external OCR-text prompt. Tables~\ref{tab:supp_sroie_and_ocr_parity_grouped} and~\ref{tab:supp_cord_clean_comparison} report the full per-field values behind the compact main-paper comparisons.

\clearpage
\makeatletter
\setlength{\@fptop}{0pt}
\setlength{\@fpsep}{6pt}
\setlength{\@fpbot}{0pt plus 1fil}
\makeatother
\begin{table}[!h]
\centering
\scriptsize
\setlength{\tabcolsep}{2pt}
\renewcommand{\arraystretch}{0.86}
\resizebox{\textwidth}{!}{%
\begin{tabular}{@{}l|l|ccccc|cccccc@{}}
\toprule
\multirow{2}{*}{Method} & \multirow{2}{*}{Input Modality} & \multicolumn{5}{c|}{SROIE Task 3 Strict F1 (\%) $\uparrow$ (KIE Domain)} & \multicolumn{6}{c}{OmniDocBench v1.5 (Original Task)} \\
& & Comp. & Date & Addr. & Total & Micro & Over.$\uparrow$ & TxtEd.$\downarrow$ & Form.$\uparrow$ & Tab.$\uparrow$ & Tab-S$\uparrow$ & RdEd.$\downarrow$ \\
\midrule

\multicolumn{13}{@{}l}{\textit{Text-Only Baselines (OCR Prompted)}} \\
GPT-4 + OCR~\cite{openai2023gpt4} & OCR Text Only & -- & -- & -- & -- & 90.60 & -- & -- & -- & -- & -- & -- \\
Llama2-7B-chat~\cite{touvron2023llama2} & OCR Text Only & -- & -- & -- & -- & 68.97 & -- & -- & -- & -- & -- & -- \\
Vicuna-1.5-7B~\cite{chiang2023vicuna} & OCR Text Only & -- & -- & -- & -- & 68.20 & -- & -- & -- & -- & -- & -- \\
Llama2 + OCR~\cite{touvron2023llama2} & OCR Text Only & -- & -- & -- & -- & 56.40 & -- & -- & -- & -- & -- & -- \\
Llama2-7B~\cite{touvron2023llama2} & OCR Text Only & -- & -- & -- & -- & 34.96 & -- & -- & -- & -- & -- & -- \\
\midrule

\multicolumn{13}{@{}l}{\textit{Spatial / Layout-Aware Baselines (No Image)}} \\
DocLLM-7B~\cite{wang2024docllm} & OCR Text + BBoxes & -- & -- & -- & -- & 91.90 & -- & -- & -- & -- & -- & -- \\
\midrule

\multicolumn{13}{@{}l}{\textit{Multimodal Baselines (Image + Text)}} \\
LayoutLLM-7B~\cite{chu2023layoutllm} & Image+Text+Layout & -- & -- & -- & -- & 72.12 & -- & -- & -- & -- & -- & -- \\
DocKylin~\cite{zhang2024dockylin} & Image + Text & -- & -- & -- & -- & 69.60 & -- & -- & -- & -- & -- & -- \\
Qwen-VL-7B~\cite{bai2023qwenvl} & Image + Text & -- & -- & -- & -- & 58.59 & -- & -- & -- & -- & -- & -- \\
LLaVA-1.5-7B~\cite{liu2023llava15} & Image + Text & -- & -- & -- & -- & 3.83 & -- & -- & -- & -- & -- & -- \\
LLaVAR-7B~\cite{zhang2023llavar} & Image + Text & -- & -- & -- & -- & 2.38 & -- & -- & -- & -- & -- & -- \\
\midrule

\multicolumn{13}{@{}l}{\textit{Base OCR Architecture}} \\
GLM-OCR-0.9B~\cite{glmocr2026} & Image & 49.76 & 78.22 & 17.81 & 74.40 & 55.04 & 94.62 & 0.040 & 93.90 & 93.96 & 96.39 & 0.044 \\
\midrule

\multicolumn{13}{@{}l}{\textit{Mixture of Task Experts Adaptation (Ours)}} \\
\rowcolor{oursrow}\textbf{\glmocrmote{} (ours)} & \textbf{Image} & \textbf{90.49} & \textbf{97.41} & \textbf{70.03} & \textbf{93.66} & \textbf{87.90} & \textbf{94.62} & \textbf{0.040} & \textbf{93.90} & \textbf{93.96} & \textbf{96.39} & \textbf{0.044} \\
\bottomrule
\end{tabular}
}
\caption{Full SROIE and OmniDocBench comparison corresponding to Table~\ref{tab:sroie_and_ocr_parity_grouped}. The first metric block evaluates receipt KIE, while the second checks preservation of the original OCR route.}
\label{tab:supp_sroie_and_ocr_parity_grouped}

\vspace{0.4em}
\setlength{\tabcolsep}{2pt}
\resizebox{\textwidth}{!}{%
\begin{tabular}{@{}l|l|ccccccccc@{}}
\toprule
\multirow{2}{*}{Method} & \multirow{2}{*}{Input Modality} & \multicolumn{9}{c}{CORD Relaxed F1 Score (\%) $\uparrow$} \\
& & Cash & Chng & Cred & Menu & Serv & SubTot & Tax & Total & Micro \\
\midrule

\multicolumn{11}{@{}l}{\textit{Text-Only Baselines (OCR Prompted)}} \\
RoBERTa~\cite{liu2019roberta} & OCR Text Only & -- & -- & -- & -- & -- & -- & -- & -- & 91.98 \\
BERT\_large/T5~\cite{devlin2019bert,raffel2020t5} & OCR Text Only & -- & -- & -- & -- & -- & -- & -- & -- & 90.25 \\
\midrule

\multicolumn{11}{@{}l}{\textit{Spatial / Layout-Aware Baselines (No Image)}} \\
LayoutLM~\cite{xu2020layoutlm} & OCR Text + BBoxes & -- & -- & -- & -- & -- & -- & -- & -- & 94.93 \\
LAMBERT~\cite{garncarek2020lambert} & OCR Text + BBoxes & -- & -- & -- & -- & -- & -- & -- & -- & 94.41 \\
\midrule

\multicolumn{11}{@{}l}{\textit{Multimodal Baselines (Image + Text + Layout)}} \\
UDOP~\cite{tang2023udop} & Image+Text+Layout & -- & -- & -- & -- & -- & -- & -- & -- & 97.60 \\
LayoutLMv2~\cite{xu2021layoutlmv2} & Image+Text+Layout & -- & -- & -- & -- & -- & -- & -- & -- & 96.01 \\
DocTr~\cite{liao2023doctr} & Image+Text+Layout & -- & -- & -- & -- & -- & -- & -- & -- & 94.40 \\
Donut~\cite{kim2022donut} & Image + Text & -- & -- & -- & -- & -- & -- & -- & -- & 91.60 \\
\midrule

\multicolumn{11}{@{}l}{\textit{Base Architecture (Before MoTE)}} \\
GLM-OCR-0.9B~\cite{glmocr2026} & Image & 11.43 & 20.59 & 16.67 & 0.00 & 47.06 & 42.55 & 15.69 & 28.12 & 20.07 \\
\midrule

\multicolumn{11}{@{}l}{\textit{Mixture of Task Experts Adaptation (Ours)}} \\
\rowcolor{oursrow}\textbf{\glmocrmote{} (ours)} & \textbf{Image} & \textbf{95.45} & \textbf{99.10} & \textbf{93.33} & \textbf{90.00} & \textbf{100.00} & \textbf{100.00} & \textbf{100.00} & \textbf{95.24} & \textbf{95.79} \\
\bottomrule
\end{tabular}
}
\caption{Full CORD KIE comparison corresponding to Table~\ref{tab:cord_clean_comparison}. \glmocrmote{} uses the receipt KIE route and predicts structured fields directly from the receipt image.}
\label{tab:supp_cord_clean_comparison}
\end{table}

\subsection{Limitations}
\label{sec:supp_limitations}

\mote{} is most effective when the task set has meaningful and stable boundaries. The prompt-conditioned selector is evaluated on generated variants of five predefined COIN intents, but this controlled result does not cover ambiguous, overlapping, or compositional requests; softer routing, hierarchical experts, or expert composition may be needed in such cases. Incremental expert addition protects existing routes by freezing old modules, but this reduces plasticity for new tasks; stronger expansion may require controlled updates to shared representations or rehearsal. 
\end{document}